\pdfoutput=1
\documentclass[sigconf]{acmart}

\usepackage{mathtools}
\usepackage{bbm}
\usepackage{subcaption}
\usepackage{multirow}
\usepackage{placeins}
\usepackage{stfloats}

\newcommand{\inst}{\boldsymbol{x}}
\newcommand{\instprime}{\boldsymbol{x'}}
\newcommand{\vone}{NHP}
\newcommand{\vthree}{HP}
\newcommand{\hypergraph}{G}
\newcommand{\nodeset}{V}
\newcommand{\edgeset}{E}
\newcommand{\embeddings}{X}
\newcommand{\nodev}{v}

\newcommand{\edgee}{\varepsilon}
\newcommand{\weightmatrix}{W}
\newcommand{\incidencematrix}{H}
\newcommand{\nodedegreematrix}{D}
\newcommand{\edgedegreematrix}{B}
\newcommand{\learnableparametermatrix}{P}
\newcommand{\perturbationrow}{\pi}
\newcommand{\perturbationmatrix}{\Pi}
\newcommand{\perturbedoperator}{S}
\newcommand{\testsplit}{V_\text{test}}
\newcommand{\indexi}{i}
\newcommand{\indexj}{j}
\newcommand{\neighborhood}{\mathcal{N}}
\newcommand{\prediction}{\hat{y}}
\newcommand{\prob}{\bold{p}}

\newcommand{\lorenzo}[1]{}
\newcommand{\fabiano}[1]{}
\newcommand{\gabri}[1]{}
\AtBeginDocument{%
  }

\setcopyright{acmlicensed}
\copyrightyear{2018}
\acmYear{2018}
\acmDOI{XXXXXXX.XXXXXXX}
\acmConference[Conference acronym 'XX]{Make sure to enter the correct
  conference title from your rights confirmation email}{June 03--05,
  2018}{Woodstock, NY}
\acmISBN{978-1-4503-XXXX-X/2018/06}

\begin{document}

\title{Counterfactual Explanations for Hypergraph Neural Networks}

\author{Fabiano Veglianti}
\orcid{0009-0007-1563-4953}
\affiliation{%
  \institution{Department of Computer Control and Management Engineering, Sapienza University}
  \city{Rome}
  \country{Italy}
}
\email{veglianti@diag.uniroma1.it}

\author{Lorenzo Antonelli}
\orcid{0009-0006-1644-0512}
\affiliation{%
  \institution{Department of Computer Control and Management Engineering, Sapienza University}
  \city{Rome}
  \country{Italy}
}
\email{antonelli@diag.uniroma1.it}

\author{Gabriele Tolomei}
\orcid{0000-0001-7471-6659}
\affiliation{%
  \institution{Department of Computer Science, Sapienza University}
  \city{Rome}
  \country{Italy}
}
\email{tolomei@di.uniroma1.it}

\renewcommand{\shortauthors}{Veglianti et al.}

\begin{abstract}
Hypergraph neural networks (HGNNs) effectively model higher-order interactions in many real-world systems but remain difficult to interpret, limiting their deployment in high-stakes settings.

We introduce CF-HyperGNNExplainer, a counterfactual explanation method for HGNNs that identifies the minimal structural changes required to alter a model's prediction. The method generates counterfactual hypergraphs using actionable edits limited to removing node-hyperedge incidences or deleting hyperedges, producing concise and structurally meaningful explanations. 
Extensive experiments on hypergraph benchmark datasets show that CF-HyperGNNExplainer generates valid and concise counterfactuals, highlighting the higher-order relations most critical to HGNN decisions.
\end{abstract}



\received{20 February 2007}
\received[revised]{12 March 2009}
\received[accepted]{5 June 2009}

\maketitle

\section{Introduction}
\fabiano{Da cambiare il formato in quello di CIKM}
Many real-world systems are governed by higher-order interactions (HOIs), in which outcomes depend on groups of entities rather than on pairwise relations alone. Examples include co-authorship teams, biochemical complexes, group communications, and multi-item user sessions, where the joint context of three or more units carries information that cannot be faithfully reduced to binary links without loss or distortion. Recent work on ``networks beyond pairwise interactions''~\cite{network-beyond-pairwise-interactions} has shown that explicitly modeling HOIs can substantially improve the ability to characterize structure, learn representations, and predict dynamical behavior in complex systems. Consequently, there is increasing interest in learning frameworks that can natively encode group interactions instead of approximating them via pairwise expansions.

Hypergraphs provide a principled representation for HOIs by allowing a hyperedge to connect an arbitrary number of nodes. Classical hypergraph learning demonstrated that hypergraphs can represent complex relationships more completely than graphs for tasks such as clustering, classification, and embedding~\cite{Learning_with_Hypergraphs}. Building on this foundation, hypergraph neural networks (HGNNs)~\cite{Hypergraph_neural_networks} and hypergraph convolutional neural networks (HGCNs)~\cite{HyperGCN} have emerged as deep learning paradigms that extend graph neural networks and graph convolutional networks to hypergraph-structured data, respectively. Reflecting this momentum, recent surveys synthesize a rapidly expanding body of deep learning on hypergraphs, exploring architectures, training strategies, and applications~\cite{A_Survey_on_Hypergraph_Neural_Networks}.

Although powerful, HGNNs suffer from limited interpretability, which reduces user trust and constrains their deployment in high-stakes settings. 
Among eXplainable AI (XAI) approaches proposed in the literature, \textit{counterfactual} (CF) explanations have attracted growing attention, as they describe minimal input changes that would lead to a different model outcome~\cite{wachter2017hjlt}.
In the graph domain, this motivation has driven substantial progress in graph neural networks (GNNs) explainability, ranging from post-hoc explainers that highlight predictive substructures (e.g., GNNExplainer~\cite{GNNExplainer} and PGExplainer~\cite{PGNNExp}) to methods that explicitly generate counterfactual graphs (e.g., CF-GNNExplainer~\cite{CF-GNNExplainer} and RCExplainer~\cite{bajaj2022robustcounterfactualexplanationsgraph}). However, transferring these ideas to hypergraphs is non-trivial: hyperedges encode higher-order relations whose combinatorial structure and semantics cannot be reduced to pairwise links without altering the meaning of the interactions. Consequently, hypergraph explainability has received comparatively limited attention. Dedicated hypergraph explainers have only emerged recently (e.g., HyperEX~\cite{HyperEX} and SHypX~\cite{shypx}), underscoring both the nascency of the area and the open need for explanation tools tailored to hypergraph learning.

In this work, we introduce CF-HyperGNNExplainer, the first counterfactual explanation technique specifically designed for hypergraph neural networks. 
Our goal is to produce valid and interpretable counterfactuals that respect the discrete, higher-order nature of hypergraphs -- e.g., by minimally editing hyperedge memberships or hyperedge existence while altering model prediction. This formulation yields explanations that answer questions of the form: ``\textit{What higher-order interactions must be removed, or rewired for the HGNN to change its decision?}''.

Our main contributions are as follows: (1) we introduce CF-HyperGNNExplainer, the first counterfactual explanation method tailored to HGNNs; 
(2) we propose two variants that operate at different granularities, i.e., removing node-hyperedge incidences (\vone) or deleting entire hyperedges (\vthree);
(3) we provide an empirical evaluation of the proposed variants on hypergraph benchmark datasets, including comparisons with graph-based counterfactual explainers and existing hypergraph-native explainers; 
(4) we release the entire source codebase to encourage reproducibility at the following URL: \url{https://anonymous.4open.science/r/CF-HyperGNNExplainer-0716}.

\section{Related Work}

\subsection{Counterfactual Explanations}\label{rel-work-counterfactuals}
Counterfactual explanations have emerged as a prominent class of instance-based explanations for AI/ML model predictions. Formally, given a classifier $f$ and an input $\inst$ for which $f(\inst) = y$, a counterfactual explanation is a perturbed version of the input $\instprime$, such that $f(\instprime) = y'$ (where $y' \neq y$) and the difference between $\inst$ and $\instprime$ is minimized according to some distance metric~\cite{wachter2017hjlt}. 

These explanation methods have been successfully applied to specific model classes (e.g., tree ensembles such as FeatureTweaking~\cite{tolomei2017kdd,tolomei2021tkde} and FOCUS~\cite{lucic2022aaai}) as well as in model-agnostic settings (e.g., DiCE~\cite{mothilal2020facct}, FACE~\cite{poyiadzi2020aies}, and ReLAX~\cite{chen2022cikm,chen2024tai}). They have also been explored across different tasks, such as recommendation~\cite{chen2025tors,scarcelli2025xai}, and types of data, particularly tabular data~\cite{movin2025tai}. More recently, they have been combined with large language models to translate technical counterfactual explanations into coherent natural-language narratives~\cite{giorgi2025aistats,vineis2026xai,silvestri2025survey}, and have also proven effective in mitigating the risk of overfitting~\cite{giorgi2026kdd}.

In the context of graph-structured data, CFs aim to answer: ``\textit{How should the graph structure or node features be minimally modified to change the model's prediction?}". A widely used method is CF-GNNExplainer~\cite{CF-GNNExplainer}, which formulates counterfactual generation as an optimization problem over the graph structure, seeking the minimal perturbation to the adjacency matrix that induces a change in the model’s prediction for a given target node. 
A key contribution of CF-GNNExplainer is its explicit focus on structural counterfactuals, emphasizing edge deletions as the primary explanation mechanism, while keeping node features fixed. This formulation enables model-agnostic applicability to a wide range of message-passing GNN architectures and provides intuitive explanations in terms of graph edits. However, the method is inherently designed for pairwise relational data represented by adjacency matrices. As a result, CF-GNNExplainer cannot be directly extended to HGNNs, where higher-order relationships are encoded through incidence structures and hyperedges connecting arbitrary-sized node sets.

\subsection{Hypergraph Neural Network Explainability}\label{rel-work-hnnexp}
Despite the rapid progress in HGNNs, explainability for hypergraph-based classifiers has only recently started to receive dedicated attention. Existing approaches largely extend the intuition of subgraph-based explanations to the hypergraph setting, aiming to identify sub-hypergraphs or node-hyperedge attributions that drive a prediction.
HyperEX~\cite{HyperEX} is a post-hoc framework that scores the importance of node-hyperedge pairs and extracts explanatory sub-hypergraphs for HGNN predictions.
SHypX~\cite{shypx} is a model-agnostic explainer that discretely samples explanation sub-hypergraphs for local attribution, and additionally derives global concepts through unsupervised extraction.
For intrinsically interpretable designs, SHy~\cite{SHy2025} introduces a self-explaining HGNN for electronic health
records based diagnosis prediction, producing distinct patient-specific temporal phenotypes as personalized explanations.
Finally, models can embed interpretability into their architecture; for instance, Bi et al.~\cite{explainable-hgnn} propose an explainable and programmable hypergraph convolutional network for imaging-genetics fusion that links decisions to hypergraph components and programmable constraints. 

Therefore, given that no counterfactual explanation methods have been developed for HGNNs and that existing graph-based approaches, such as CF-GNNExplainer, cannot be directly extended to hypergraphs, in this work, we propose CF-HyperGNNExplainer, the first counterfactual explanation method specifically designed for hypergraph neural networks.
\section{Background}
Our approach builds on the core idea behind CF-GNNExplainer~\cite{CF-GNNExplainer}: generate counterfactual explanations by searching for the smallest structural edit to the input relational structure that flips a trained model’s prediction.
For this reason, in this section we review CF-GNNExplainer, to clarify the counterfactual optimization template we inherit, and the hypergraph neural network propagation operator, to make explicit where and how structural edits can be injected in a differentiable manner.

\noindent \textbf{\textit{Notation.}} We adopt the following conventions throughout the paper. Vectors are bold lowercase (e.g., $\inst, \boldsymbol{y}\in\mathbb{R}^d$). Matrices are uppercase roman (e.g., $A,\edgedegreematrix\in\mathbb{R}^{n\times m}$). The $(\indexi,\indexj)$-th entry of $A$ is $A_{ij}$. The $\indexi$-th row of $A$ is $A_{\indexi.}\in\mathbb{R}^{1\times m}$ and the $\indexj$-th column is $A_{.\indexj}\in\mathbb{R}^{n\times 1}$. The identity matrix is $I_d$, the all-zeros vector is $\mathbf{0}$, and the all-ones vector is $\mathbf{1}$.


\subsection{CF-GNNExplainer}
Given a target node $\nodev$ and its $k$-hop computational subgraph $G_v = (A, \embeddings)$, where $A$ is the adjacency matrix and $\embeddings$ the feature matrix, the method search for a perturbation matrix $P$ that minimizes the following objective function:
\begin{equation*}
\mathcal{L} = -\mathcal{L}_{pred}(f(A \odot P, \embeddings), \prediction) + \beta \mathcal{L}_{dist}(P)
\end{equation*}
Here, $\odot$ denotes the element-wise product, hence $A \odot P$ is the counterfactual adjacency matrix. The first term, $\mathcal{L}_{pred}$, encourages the model to flip the prediction from the original prediction $y$. The second term, $\mathcal{L}_{dist}$, acts as a regularizer to ensure that the resulting counterfactual remains ``close'' to the original graph. To handle the discrete nature of edges, $P$ is typically relaxed to continuous values during optimization and thresholded afterward to produce a binary graph structure.

We inherit from CF-GNNExplainer the underlying idea; however, we adapt it to HGNNs. Since CF-GNNExplainer cannot be trivially generalized to HGNNs, we end up with two different algorithms to produce HGNNs' counterfactual explanations, detailed in Section~\ref{sec: problem formulation}. 

\subsection{Hypergraph Neural Networks}
\label{subsec:hypergraph}

Let $\hypergraph=(\nodeset,\edgeset)$ be a hypergraph with $|\nodeset| = N$ nodes and $|\edgeset| = M$ hyperedges.
Each hyperedge $\edgee \in \edgeset$ has an associated positive weight $W_{\edgee\edgee}$, and all
hyperedge weights are stored in the diagonal matrix $W \in \mathbb{R}^{M\times M}$.
Unlike simple graphs, a hypergraph is commonly represented by its incidence matrix
$\incidencematrix \in \mathbb{R}^{N\times M}$, where
\begin{equation*}
H_{\indexi\edgee} =
\begin{cases}
1, & \text{if } \nodev_i \in \nodeset \text{ is incident to hyperedge } \edgee \in \edgeset,\\
0, & \text{otherwise.}
\end{cases}
\end{equation*}
The node degree matrix $\nodedegreematrix \in \mathbb{R}^{N\times N}$ and the hyperedge degree matrix
$\edgedegreematrix \in \mathbb{R}^{M\times M}$ are diagonal matrices defined by
\begin{equation}
D_{ii}=\sum_{\edgee=1}^{M} W_{\edgee\edgee} H_{\indexi\edgee},\quad B_{\edgee\edgee}=\sum_{\indexi=1}^{N} H_{\indexi\edgee}.
\label{eq:node-edge-degree}
\end{equation}

\smallskip
\noindent \textbf{\textit{Convolution on Hypergraphs.}}
Information propagation on hypergraphs assumes: (\indexi) nodes sharing a hyperedge should
exchange more information, and (ii) heavier hyperedges should contribute more to the
propagation. With these principles, one layer of hypergraph convolution updates the
embedding of node $\indexi$ via
\begin{equation}
\inst^{(l+1)}_i=\sigma\!\left(\sum_{\indexj=1}^{N}\sum_{\edgee=1}^{M}
H_{\indexi\edgee}H_{\indexj\edgee}W_{\edgee\edgee}\, \inst^{(l)}_{\indexj}\, \learnableparametermatrix^{(l)}\right),
\label{eq:nodewise-update}
\end{equation}
where $\inst^{(l)}_i$ is the $\indexi$-th node embedding at layer $l$, $\sigma(\cdot)$ is a nonlinearity
(e.g., LeakyReLU or ELU), and $\learnableparametermatrix^{(l)}\in\mathbb{R}^{F^{(l)}\times F^{(l+1)}}$ is the learnable weight
matrix at layer $l$. In matrix form,
\begin{equation*}
\embeddings^{(l+1)}=\sigma\!\left(HWH^{\top} \embeddings^{(l)} \learnableparametermatrix^{(l)}\right),
\label{eq:matrix-unormalized}
\end{equation*}
with $\embeddings^{(l)}\in\mathbb{R}^{N\times F^{(l)}}$ and $\embeddings^{(l+1)}\in\mathbb{R}^{N\times F^{(l+1)}}$.

\noindent \textbf{\textit{Normalization.}}
The operator $HWH^{\top}$ does not control the spectral radius, so stacking many layers
can cause scaling issues and unstable optimization. A symmetric normalization yields
the normalized propagation
\begin{equation}
\embeddings^{(l+1)}=\sigma\!\left(\nodedegreematrix^{-\frac{1}{2}} \incidencematrix \weightmatrix \edgedegreematrix^{-1} \incidencematrix^{\top} \nodedegreematrix^{-\frac{1}{2}} \embeddings^{(l)} \learnableparametermatrix^{(l)}\right).
\label{eq:symm-norm}
\end{equation}
Here $\nodedegreematrix$ and $\edgedegreematrix$ are the degree matrices in \eqref{eq:node-edge-degree}.
One can show the largest eigenvalue of $\nodedegreematrix^{-\frac{1}{2}} \incidencematrix \weightmatrix \edgedegreematrix^{-1} \incidencematrix^{\top} \nodedegreematrix^{-\frac{1}{2}}$ is at
most $1$ ensuring stability during optimization.

\smallskip
\noindent \textbf{\textit{Node Interaction.}}
Hypergraph convolution respects the neighborhood of each node: a node only aggregates signals from nodes that share at least one hyperedge with it. Let consider
\begin{equation*}
\incidencematrix \incidencematrix^{\top}\in\mathbb{R}^{N\times N},
\qquad
(\incidencematrix \incidencematrix^{\top})_{ij} \;=\; \sum_{\edgee=1}^{M} \incidencematrix_{\indexi\edgee}\incidencematrix_{\indexj\edgee},
\label{eq:coincidence}
\end{equation*}
be the (unweighted) node co-incidence matrix that counts how many hyperedges make $v_i$ and $v_j$ co-incident.
The (closed) neighborhood of $\indexi$ is then
\begin{equation*}
\neighborhood(\indexi)\;=\;\{\, \indexj\in\{1,\dots,N\}\;:\; (\incidencematrix \incidencematrix^{\top})_{ij}>0 \,\}.
\label{eq:neighborhood}
\end{equation*}

With this notation, the node-wise update in equation~\eqref{eq:nodewise-update} can be written as
\begin{equation}
x^{(l+1)}_i
\;=\;
\sigma\!\left(
\sum_{\indexj\in\neighborhood(\indexi)} \bigg(\sum_{\edgee=1}^{M}
\!\!\incidencematrix_{\indexi\edgee}\incidencematrix_{\indexj\edgee}\weightmatrix_{\edgee\edgee}\bigg)
\, x^{(l)}_{\indexj}\, \learnableparametermatrix^{(l)} \right),
\label{eq:nodewise-neighborhood}
\end{equation}
which makes explicit that only $\indexj\in\neighborhood(\indexi)$ contribute to $x^{(l+1)}_i$.
In matrix form, the normalized propagation used in equation~\eqref{eq:symm-norm} reads
\begin{equation*}
\begin{aligned}
    &\embeddings^{(l+1)} \;=\; \sigma\!\left(\perturbedoperator\,\embeddings^{(l)} \learnableparametermatrix^{(l)}\right),\\
&\perturbedoperator \;\coloneqq\; \nodedegreematrix^{-\frac{1}{2}} \incidencematrix \weightmatrix \edgedegreematrix^{-1} \incidencematrix^{\top} \nodedegreematrix^{-\frac{1}{2}},\\
\end{aligned}
\end{equation*}
\label{eq:normalized-operator}
where $\perturbedoperator$ inherits the same sparsity pattern of $HH^{\top}$, and thus preserves the neighborhood constraint of Equation~\eqref{eq:nodewise-neighborhood}.

Similarly to what happens with graphs, after n layers, a node’s representation depends on nodes within n hops in the node–hyperedge incidence, so we define the n-hop neighborhood for node $\nodev_{\indexi}$ as:
\begin{equation*}
\neighborhood^{n}(\indexi)\;=\;\{\, \indexj\in\{1,\dots,N\}\;:\; ((\incidencematrix \incidencematrix^{\top})^n)_{ij}>0 \,\}.
\label{eq:n-neighborhood}
\end{equation*}

\smallskip
\noindent \textbf{\textit{Model.}}
Since $\embeddings^{(l+1)}$ is differentiable w.r.t.\ both $\embeddings^{(l)}$ and $\learnableparametermatrix^{(l)}$, hypergraph convolution
layers can be trained end-to-end via gradient-based optimization.

Let $\embeddings = \embeddings^{(0)}\in\mathbb{R}^{N\times F^{(0)}}$ be the input node features. A hypergraph neural network is obtained by stacking the $L$ convolution layers in~\eqref{eq:symm-norm}, producing $\embeddings^{(L)}$. A node-level predictor is then defined as
\begin{equation*}
\prediction \;=\; f(\incidencematrix,\embeddings, \learnableparametermatrix^{(l)})\;\coloneqq\;\text{softmax}\!\big(\embeddings^{(L)}\learnableparametermatrix^{(L)}\big)\in\mathbb{R}^{N\times C},
\label{eq:prediction-model}
\end{equation*}
where $C$ is the number of classes in our application and $\learnableparametermatrix^{(l)}$, with $l=0,...,L$, collect all learnable parameters. In the following, we will omit the dependency from the learnable parameters $\learnableparametermatrix$ since the counterfactual explanation generation problem assume the model parameters to be frozen.
We focus on node classification task: given labeled nodes, the model is trained to predict these labels. 
\section{Problem Formualtion}\label{sec: problem formulation}
\subsection{Perturbed Signal Propagation}

A convolutional layer on a hypergraph depends on: (\indexi) the structure encoded by the incidence matrix $\incidencematrix$ (which determines the node neighborhood), (ii) the current node embeddings $\embeddings^{(l)}$, and (iii) the layer parameters $\learnableparametermatrix^{(l)}$.
In analogy with counterfactual sparsification for graphs---where the forward map depends on the adjacency $A_v$, the local features $\embeddings_v$, and the weights, resulting in $f(A_v, X_v; W)$, and a learnable perturbation P is introduced to obtain a perturbed map $g(A_v, \embeddings_v, W; P)$---we introduce a \emph{learnable perturbation matrix} $\perturbationmatrix$ that modulates the neighborhood used by the layer.

Formally, we define a perturbed propagation
\begin{equation}
\embeddings^{(l+1)} \;=\; \sigma\!\left(\perturbedoperator(\perturbationmatrix)\,\embeddings^{(l)} \learnableparametermatrix^{(l)}\right),
\label{eq:cf-general}
\end{equation}
where the operator $\perturbedoperator(\perturbationmatrix)$ is obtained by applying the perturbation matrix $\perturbationmatrix$ (element-wise, i.e., Hadamard-wise) to the incidence matrix $\incidencematrix$, thus resulting in:
\begin{equation}
    \perturbedoperator(\perturbationmatrix) \;\coloneqq\; \nodedegreematrix^{-\frac{1}{2}} (\incidencematrix \odot \perturbationmatrix) \weightmatrix \edgedegreematrix^{-1} (\incidencematrix \odot \perturbationmatrix)^{\top} \nodedegreematrix^{-\frac{1}{2}},
    \label{eq:s-perturb}
\end{equation}
with the degrees being recomputed whenever the incidence is modified. This mirrors the CF-GNNExplainer's idea of inserting a learnable, differentiable mask into the neighborhood aggregation so that the logits depend on $(\incidencematrix, \embeddings^{(l)}, \learnableparametermatrix)$ and on $\perturbationmatrix$.

\subsection{Loss Formulation}
\label{sec:loss_formulation}

Similarly to CF-GNNExplainer, we learn $\Pi$ by solving an optimization problem, where the
HGNN parameters are kept fixed and only the mask is optimized. 

\noindent \textbf{\textit{Optimization Objective.}}
Let $f$ denote the trained HGNN and let $S(\Pi)$ be the perturbed propagation operator introduced in equation~\eqref{eq:s-perturb}.
Denote by $\prediction$ the original predicted label for the explained node under the unperturbed hypergraph, we then learn $\Pi$ by minimizing the following loss:
\begin{equation}
\mathcal{L}(\Pi)
=
-\mathcal{L}_{\text{pred}}\!\big(f\big((\incidencematrix \odot \perturbationmatrix ), \embeddings\big),\, \prediction\big)
\;+\;
\beta\,\mathcal{L}_{\text{dist}}(\Pi),
\label{eq:loss_compact_hyper}
\end{equation}
where $-\mathcal{L}_{\text{pred}}$ penalizes the model's confidence in the original class $\prediction$ under the
perturbed propagation (thus encouraging a prediction change), while $\mathcal{L}_{\text{dist}}$ enforces
minimality of the edit and $\beta>0$ controls the trade-off.

To make equation~\eqref{eq:loss_compact_hyper} explicit, we instantiate $-\mathcal{L}_{\text{pred}}$ as the log-probability
assigned to the original class $prediction$ by the perturbed model output so that minimizing $-\mathcal{L}_{\text{pred}}$ the perturbed prediction is pushed away
from the original decision without specifying a target class. The regularization term $\mathcal{L}_{\text{dist}}$ is instantiated as the L1 distance between the perturbation matrix and the original incidence matrix. Finally, in direct analogy with CF-GNNExplainer,
the prediction term is made active only while the prediction has not yet flipped by multiplying it by the following term
$\mathbbm{1}[ \arg\max f(\incidencematrix \odot \perturbationmatrix ,\embeddings)=\prediction ]$, with $\mathbbm{1}(\cdot)$ being the indicator function.

\noindent \textbf{\textit{Continuous Relaxation and Thresholding.}}
Since the incidence matrix $H$ is discrete, the desired perturbation is discrete as well.
However, to enable gradient-based optimization, we optimize a continuous relaxation of the mask.
Specifically, we learn a continuous perturbation matrix, we map it to $(0,1)$ via a sigmoid, and use the resulting soft mask in the forward pass.
At evaluation time, we obtain an actionable counterfactual hypergraph by thresholding the relaxed mask, using $0.5$ as threshold.

\subsection{Method Variants}
\label{subsec:method-variants}

We consider two variants that differ in how the $\perturbationmatrix$ is initialized.

\textbf{\textit{(\vone) Node–Hyperedge Perturbation.}}
\emph{Goal: eliminate the participation of a specific node $\indexi$ in selected incident hyperedges $\edgee$.}

Let $\boldsymbol{\perturbationrow}^{(\indexi)}\!\in[0,1]^M$ be a vector of hyperedge-wise learnable perturbation coefficients for node $\indexi$, and lift it to a rectangular mask
$\perturbationmatrix^{(\indexi)}\in[0,1]^{N\times M}$ by
\[
(\perturbationmatrix^{(\indexi)})_{k\edgee}
\;=\;
\begin{cases}
\perturbationrow^{(\indexi)}_{\edgee}, & \text{if}\,\,k = \indexi,\\[2pt]
1, & \text{if}\,\,k \neq \indexi.
\end{cases}
\]
For any neighbor $\indexj$ that communicates with $\indexi$ through a hyperedge $\edgee$, the message from $\indexj$ to $\indexi$ now carries a factor
$\perturbationrow^{(\indexi)}_{\edgee}$ (since $\incidencematrix'_{\indexi\edgee} = \perturbationrow^{(\indexi)}_{\edgee} \incidencematrix_{\indexi\edgee}$ while $\incidencematrix'_{\indexj\edgee}=\incidencematrix_{\indexj\edgee}$), so that setting $\perturbationrow^{(\indexi)}_{\edgee}\!\approx 0$ softly (or, in the binary case, hard-) removes the contribution carried through $\edgee$ to the representation of $v_i$ without altering the incidences of other nodes.

\textbf{\textit{(\vthree) Hyperedge Perturbation.}}
\emph{Goal: eliminate entire hyperedges in the local neighborhood of a target node $\nodev_{\indexi}$.}
Let $\perturbationmatrix\!\in[0,1]^{N\times M}$ be row-constant and tunable only in the n-hop neighborhood of $\nodev_{\indexi}$, i.e.,
\[
\forall k=1,...,N\quad
(\perturbationmatrix^{(\indexi)})_{k\edgee}
\;=\;
\begin{cases}
\perturbationrow_{\edgee}, & \text{if}\,\,\edgee \in \neighborhood^n(i),\\[2pt]
1, & \text{if}\,\,\edgee \notin \neighborhood^n(i).
\end{cases}
\]
For every hyperedge $\edgee\in\neighborhood^n(i)$\footnote{With a slight abuse of notation, we here identify a neighborhood with the subhypergraph induced by its nodes. Thus, we say that an hyperedge belongs to the neighborhood if it belongs to the corresponding induced subhypergraph.} and every incident node $k$ to it we have
$\incidencematrix'_{k\edgee}=\perturbationrow_{\edgee}\incidencematrix_{k\edgee}$, while hyperedges outside $\neighborhood^n(i)$ are unchanged.
Hence, any message that traverses a hyperedge $\edgee$ within the $n$-hop computation around $\nodev_{\indexi}$ is uniformly scaled by $\perturbationrow_{\edgee}$ (note that $\edgee$ need not be incident to $\nodev_{\indexi}$, it only needs to lie on an information-flow path in $\neighborhood^n(i)$).
Setting $\perturbationrow_{\edgee}\approx 0$ therefore suppresses all interactions mediated by $\edgee$ in that neighborhood, effectively removing the corresponding conduit of information flow for all nodes participating in $\edgee$.

\paragraph{Remarks.}
All variants plug into equation~\eqref{eq:cf-general} by treating the perturbation as a differentiable mask over the neighborhood structure.
When the incidence matrix is modified (\vone, \vthree), we recompute the induced node and hyperedge degrees so that normalization remains consistent with the perturbed connectivity.
The two variants offer complementary control: (\vthree) provides hyperedge-level gating, while (\vone) enables finer node-hyperedge edits, allowing different granularities of counterfactual neighborhood intervention.

\begin{figure}
    \centering
    \includegraphics[width=\linewidth]{nhp_hp_figure.pdf}
    \caption{The two variants of CF-HyperGNNExplainer. (NHP) removes the incidence between the target node and a selected hyperedge. (HP) removes an entire hyperedge, suppressing all interactions. Dashed red circles highlight the target nodes for which the counterfactuals are generated.}
    \label{fig:method_variants}
\end{figure}

\section{Experimental Setup}
\subsection{Implementation Details}\label{subsec: implementation-details}
We run all experiments on a machine equipped with an AMD Ryzen 9 7900 12-Core Processor and an NVIDIA GeForce RTX 4090 GPU. Our implementation is based on the PyTorch Geometric framework. 



We implemented our method using sparse matrices with the COO format. The COO format stores only the non-zero elements of a matrix along with their coordinates, which is particularly efficient for the hypergraph incidence matrices, which are typically very sparse. The code is publicly available on GitHub\footnote{\url{https://anonymous.4open.science/r/CF-HyperGNNExplainer-0716/README.md}}.

\subsection{Datasets and Models}
We assess the performance of CF-HyperGNNExplainer on a widely used hypergraph-native datasets collection from AllSet~\cite{allset}. This collection is made of readaptation of 13 benchmark datasets: cocitation networks Cora, Citeseer and Pubmed~\cite{HyperGCN}, coauthorship Cora and DBLP~\cite{HyperGCN}, 20Newsgroups, Mushroom and Zoo from UCI Repository~\cite{UCI}, the Princeton CAD ModelNet40~\cite{ModelNet40}, the NTU2012 3D dataset~\cite{NTU2012} and the Yelp, House~\cite{HouseDataset} and Walmart~\cite{WalmartDataset}.
Details on the datasets are shown in table~\ref{tab:dataset-statistics}. Our work excludes \texttt{20Newsgroups}, \texttt{Walmart}, and \texttt{Yelp} as they exceeded available memory capacity of our hardware.

\begin{table*}[t]
\caption{Summary statistics of the benchmark hypergraph datasets used in the experimental evaluation.}
\label{tab:dataset-statistics}
\centering
\scriptsize
\resizebox{\textwidth}{!}{%
\begin{tabular}{lrrrrrrrrrrrrr}
\toprule
\textbf{Feature} & \textbf{Cora} & \textbf{Citeseet} & \textbf{Pubmed} & \textbf{Cora-CA} & \textbf{DBLP-CA} & \textbf{Zoo} & \textbf{20News} & \textbf{Mushroom} & \textbf{NTU2012} & \textbf{ModelNet40} & \textbf{Yelp} & \textbf{House} & \textbf{Walmart} \\
\midrule
\texttt{$|V|$} & 2708 & 3312 & 19717 & 2708 & 41302 & 101 & 16242 & 8124 & 2012 & 12311 & 50758 & 1290 & 88860 \\
\texttt{$|E|$} & 1579 & 1079 & 7963 & 1072 & 22363 & 43 & 100 & 298 & 2012 & 12311 & 679302 & 341 & 69906 \\
\texttt{$F^{(0)}$} & 1433 & 3703 & 500 & 1433 & 1425 & 16 & 100 & 22 & 100 & 100 & 1862 & 6 & 6 \\
\texttt{$\#classes$} & 7 & 6 & 3 & 7 & 6 & 7 & 4 & 2 & 67 & 40 & 9 & 2 & 11 \\
\texttt{$H_{\text{nnz}}$} & 4786 & 3453 & 34629 & 4585 & 99561 & 1717 & 65451 & 40620 & 10060 & 61555 & 4523594 & 11841 & 460629 \\
\texttt{$min(D_{ii})$} & 0 & 0 & 0 & 0 & 1 & 17 & 1 & 5 & 1 & 1 & 1 & 0 & 0 \\
\texttt{$max(D_{ii})$} & 145 & 88 & 99 & 23 & 18 & 17 & 44 & 5 & 19 & 30 & 7855 & 44 & 5733 \\
\texttt{$median(D_{ii})$} & 1 & 0 & 0 & 2 & 2 & 17 & 3 & 5 & 5 & 4 & 35 & 7 & 2 \\
\texttt{$avg(D_{ii})$} & 1.76736 & 1.04257 & 1.7563 & 1.69313 & 2.41056 & 17 & 4.02974 & 5 & 5 & 5 & 89.1208 & 9.17907 & 5.18376 \\
\texttt{$min(B_{\epsilon\epsilon})$} & 2 & 2 & 2 & 2 & 2 & 1 & 29 & 1 & 5 & 5 & 2 & 1 & 2 \\
\texttt{$max(B_{\epsilon\epsilon})$} & 5 & 26 & 171 & 43 & 202 & 93 & 2241 & 1808 & 5 & 5 & 2838 & 81 & 25 \\
\texttt{$median(B_{\epsilon\epsilon})$} & 3 & 2 & 3 & 3 & 3 & 40 & 537 & 72 & 5 & 5 & 3 & 40 & 5 \\
\texttt{$avg(B_{\epsilon\epsilon})$} & 3.03103 & 3.20019 & 4.34874 & 4.27705 & 4.45204 & 39.9302 & 654.51 & 136.309 & 5 & 5 & 6.65918 & 34.7243 & 6.58926 \\
\bottomrule
\end{tabular}%
}
\end{table*}

The model used in all the experiments is a 3-layer Hypergraph Convolutional Network inspired by the architecture of the one used by Lucic et al.~\cite{CF-GNNExplainer}, with hidden dimensions of 64 and 32, Leaky ReLU activation function, a dropout probability of 0.5 used during training, and a final linear classification layer. The model was trained for 200 epochs using a learning rate of 0.01, a weight decay of $5 \times 10^{-4}$ and SGD as optimizer.

\subsection{Baselines}
We compare CF-HyperGNNExplainer against two categories of baselines.


First, we compare our method with CF-GNNExplainer~\cite{CF-GNNExplainer}. This is the natural comparison when focusing on counterfactual explanations: since no existing method natively generates counterfactual explanations for hypergraphs, the only viable approach is to first convert each hypergraph into a graph and then apply a graph-native counterfactual generation method. For the conversion, we adopted star expansion, where each hyperedge $\epsilon$ is replaced by an auxiliary node $v_{\epsilon}$, and all nodes incident to $\epsilon$ are connected to $v_{\epsilon}$, as this expansion do not let any information vanish. Among counterfactual generation methods for graph data, we choose CF-GNNExplainer because it is the closest to our setting: as both search for minimal structural perturbations based on edge removal.

Second, we consider existing explainability methods specifically designed for hypergraphs, HyperEX~\cite{HyperEX} and SHypX~\cite{shypx}. These methods operate directly on hypergraph-structured data and therefore provide the most natural comparison for assessing the quality of explanations produced by our method. However, unlike CF-HyperGNNExplainer, they are not designed to generate counterfactual explanations through minimal structural interventions. In contrast, their objective is the opposite to ours: they aim to identify the smallest subhypergraph that preserves the model's prediction, whereas CF-HyperGNNExplainer aims to identify the smallest structural edit that alters it.

\subsection{Metrics}\label{subsec: experiments-metrics}
For our experiments we use two different set of metrics that allow to compare our method with the chosen baselines.

For the comparison with CF-GNNExplainer, we use the same metrics used in their work, except that we apply them to hypergraphs; these metrics are reported below. Additionally to the metrics defined below, they add an \textit{accuracy} metric, which measure the proportion of counterfactual explanation that are ``correct'', meaning that the counterfactual difference, $\incidencematrix - \incidencematrix'$ in our case, is truly responsible for label flipping. In their setting, which relies on synthetic datasets this correctness can be investigated; however, on real dataset, it's impossible to investigate if the explanation is removing edges (or hyperedges) due to spurious correlation or it is not, so this metric cannot be computed and will be skipped in this comparison.


\noindent \textbf{\textit{Fidelity}} is defined as the proportion of nodes where the original predictions match the prediction for the explanations. Formally:
\begin{equation*}
\text{Fidelity} = \frac{1}{|\testsplit|} \sum_{\nodev \in \testsplit} \mathbbm{1} [f_{\nodev}(\incidencematrix', X) = f_{\nodev}(\incidencematrix,X) ]
\end{equation*}

Due to absence of \textit{accuracy}, we prefer to measure the \textbf{success} \textbf{rate} defined as follows:
$$
\text{Success Rate} = 1 - \text{Fidelity}
$$

\noindent \textbf{\textit{Explanation Size}} measures the amount of structural change required to obtain a CF example. The definition differs depending on the method's variant.
For variant \vone{}, is measured as the number of node-hyperedge incidences removed from the original incidence matrix $\incidencematrix$ for the node we are considering. For variant \vthree{}, the explanation size is measured as the number of hyperedges removed from $\incidencematrix$. Formally, the metric for variant \vone{} is:
\begin{equation*}
\text{Size}_{\vone{}} (\incidencematrix, \incidencematrix') = \sum_{i=1}^{N}\sum_{\epsilon = 1}^{M} |\incidencematrix_{i\epsilon} - \incidencematrix'_{i\epsilon}|
\end{equation*}
while, the metric for variant \vthree{} is:

\begin{equation*}
\text{Size}_{\vthree{}} = \sum_{\epsilon \in \mathcal{E}} \mathbbm{1} [\epsilon \notin \mathcal{E}_{\text{cf}}]
\end{equation*}
where $\edgeset$ is the set of hyperedges in the original hypergraph and $\edgeset_\text{cf}$ is the set of hyperedges in the CF hypergraph.

\noindent \textbf{\textit{Sparsity}} measures the fraction of the original hypergraph structure preserved in the CF explanation. For variant \vone{}, it is defined as the fraction of incidences that are preserved in the CF hypergraph's incidence matrix $\incidencematrix$. Formally:
\begin{equation*}
\text{Sparsity}_{\vone{}} = 1-\frac{\text{Size}_{\vone{}}}{\sum_{i,\epsilon} \incidencematrix_{i\epsilon}}
\end{equation*}
For variant \vthree{}, sparsity is defined as the fraction of hyperedges preserved. Formally:
\begin{equation*}
\text{Sparsity}_{\vthree{}} = 1-\frac{\text{Size}_{\vthree{}}}{|\edgeset|}
\end{equation*}

For the comparison with HyperEX and SHypX, we adhere to the metrics defined in the latter. To simplify the notation, we use $y_{\nodev} = f_{\nodev}(\incidencematrix,X)$ as the original prediction on node $\nodev$, $y'_{\nodev} = f_{\nodev}(\incidencematrix', X)$ as the prediction on node $\nodev$ given the counterfactual graph, and $\bar{y}'_{\nodev} = f_{\nodev}(\incidencematrix - \incidencematrix', X)$ as the prediction on node $\nodev$ given the counterfactual difference. Additionally use $\prob_{\nodev}^{\incidencematrix}$ as the original predicted probabilities over classes for node $\nodev$, $\prob_{\nodev}^{\incidencematrix'}$ as the predicted probabilities over classes for node $\nodev$ given the counterfactual graph, and $\prob_{\nodev}^{\incidencematrix - \incidencematrix'}$ as the predicted probabilities over classes for node $\nodev$ given the counterfactual difference.

\noindent \textbf{\textit{Accuracy Fidelity}}
measures the fraction of test nodes whose prediction changes under the counterfactual hypergraph ($\mathrm{Fid}_{+}^{\text{acc}}$), or under the counterfactual difference hypergraph ($\mathrm{Fid}_{-}^{\text{acc}}$.
\begin{equation*}
\begin{aligned}
&\mathrm{Fid}_{-}^{\text{acc}} =
\frac{1}{|V_{\text{test}}|} \sum_{v\in V_{\text{test}}} \mathbbm{1}(\bar{y}'_v \neq y_v), \\
&\mathrm{Fid}_{+}^{\text{acc}} = 
\frac{1}{|V_{\text{test}}|} \sum_{v\in V_{\text{test}}} \mathbbm{1}(y'_v \neq y_v),    \\
\end{aligned}
\end{equation*}

\noindent \textbf{\textit{Kullback-Leibler Divergence Fidelity}}
measures the divergence between the original prediction distribution and the one on the counterfactual hypergraph ($\mathrm{Fid}_{+}^{\text{KL}}$) or the counterfactual difference ($\mathrm{Fid}_{-}^{\text{KL}}$) hypergraph.
\begin{equation*}
\begin{aligned}
&\mathrm{Fid}_{-}^{\text{KL}} =
\frac{1}{|V_{\text{test}}|} \sum_{v\in V_{\text{test}}} \sum_{c \in C} \prob_{\nodev}^{\incidencematrix - \incidencematrix'}(c) \log\biggr(\frac{\prob_{\nodev}^{\incidencematrix - \incidencematrix'}(c)}{\prob_{\nodev}^{\incidencematrix}(c)}\biggr), \\
&\mathrm{Fid}_{+}^{\text{KL}} = 
\frac{1}{|V_{\text{test}}|} \sum_{v\in V_{\text{test}}} \sum_{c \in C} \prob_{\nodev}^{\incidencematrix'}(c) \log\biggr(\frac{\prob_{\nodev}^{\incidencematrix'}(c)}{\prob_{\nodev}^{\incidencematrix}(c)}\biggr),    \\
\end{aligned}
\end{equation*}

\noindent \textbf{\textit{Total Variation Fidelity}}
measures the total variation distance between the original prediction distribution and the one on the counterfactual hypergraph ($\mathrm{Fid}_{+}^{\text{TV}}$) or the counterfactual difference ($\mathrm{Fid}_{-}^{\text{TV}}$) hypergraph.
\begin{equation*}
\begin{aligned}
&\mathrm{Fid}_{-}^{\text{TV}} =
\frac{1}{|V_{\text{test}}|} \sum_{v\in V_{\text{test}}} \frac{1}{2}\sum_{c \in C} \bigr|\prob_{\nodev}^{\incidencematrix - \incidencematrix'}(c) - \prob_{\nodev}^{\incidencematrix}(c)\bigr|, \\
&\mathrm{Fid}_{+}^{\text{TV}} = 
\frac{1}{|V_{\text{test}}|} \sum_{v\in V_{\text{test}}} \frac{1}{2}\sum_{c \in C} \bigr|\prob_{\nodev}^{\incidencematrix'}(c) - \prob_{\nodev}^{\incidencematrix}(c)\bigr|,    \\
\end{aligned}
\end{equation*}

\noindent \textbf{\textit{Xent Fidelity}}
measures fidelity through the cross-entropy between the original prediction distribution and the counterfactual one ($\mathrm{Fid}_{+}^{\text{Xent}}$) or the counterfactual difference hypergraph ($\mathrm{Fid}_{-}^{\text{Xent}}$).
\begin{equation*}
\begin{aligned}
&\mathrm{Fid}_{-}^{\text{Xent}} =
-\frac{1}{|V_{\text{test}}|} \sum_{v\in V_{\text{test}}} \sum_{c \in C} \prob_{\nodev}^{\incidencematrix}(c) \log\bigr(\prob_{\nodev}^{\incidencematrix - \incidencematrix'}(c)\bigr), \\
&\mathrm{Fid}_{+}^{\text{Xent}} = 
-\frac{1}{|V_{\text{test}}|} \sum_{v\in V_{\text{test}}} \sum_{c \in C} \prob_{\nodev}^{\incidencematrix}(c) \log\bigr(\prob_{\nodev}^{\incidencematrix'}(c)\bigr),    \\
\end{aligned}
\end{equation*}

\noindent \textbf{\textit{Density}} measures the fraction of the original hypergraph structure removed in the CF explanation. Formally:
$$ Density = 1-Sparsity.$$

\subsection{Hyperparameter Search}\label{subsec: hyperparameter-search}
Our method contains one hyperparameter, the regularization parameter $\beta$. This parameter in eq. \eqref{eq:loss_compact_hyper} offer a trade-off between minimality of the edit and prediction change in the optimization objective function. Following original work on counterfactual explanations~\cite{wachter2017hjlt}, we select the largest value of $\beta$ that still ensures a prediction flip.

Furthermore, we measured the effect of different optimization hyperparameters for CF-HyperGNNExplainer, focusing on learning rate and epochs of the SGD optimizer. In particular, we tested learning rates $\eta \in \{1, 10,100\}$ and epochs value in $\{200, 400, 600\}$ on the Cora-CA dataset.

\noindent \textbf{\textit{Adaptive Learning Rate.}} Accoding to \cite{giorgi2026kdd} and our observation, CF generation is harder for instances where model is overly confident since this result in flat loss landscape that would require higher learning rate or much more epochs. To avoid this issue, we implement a dynamic learning rate selection algorithm which chooses the learning rate value taking into account the remaining epochs and the loss landscape with the following formula:
\begin{equation}\label{eq: learning-rate-dynamic}
        \eta = \frac{(\tau+ \log C)/T_r}
    {\|\nabla_\Pi(-\log \prob_{\Pi_0}(y)\|_2^2 + \varepsilon},
\end{equation}
with $-\log \prob_{\Pi_0}(y)$ being the log-probability assigned to the original class prediction under the perturbation matrix $\Pi_0$, $T_r$ being the number of remaining epochs, $\tau = 0.1$ being a tolerance and $\varepsilon>0$ a small positive value to avoid division by 0. We implement three variants of this learning rate selection algorithm: one computes the learning rate at the very first epoch (\texttt{1E}), one updates it every power of two epochs (\texttt{Po2}) and one updates it every epoch (\texttt{EE}). The three variants offer a trade off between adaptability and overhead. The rationale behind this formula is clear in the third variant: at the end of the counterfactual generation it allow the log-probability assigned to the original class prediction to approximatively become $\tau + \log C$ which guarantees the original class to be no more the most probable class, so it guarantees the label flipping. 

\section{Results}
In this section, we present an evaluation of CF-HyperGNNExplainer on multiple benchmark datasets. We first compare our approach with existing graph-based CF explanation methods, then we compare our approach with hypergraph-based explanation methods proposed in literature. Finally, we provide an ablation study to analyze the impact of key optimization hyperparameters.


As anticipated in Section~\ref{subsec: implementation-details},  \texttt{20Newsgroups}, \texttt{Walmart}, and \texttt{Yelp} datasets could not be included in the experiments as they exceeded available memory capacity.

\subsection{Hyperparameter Calibration}

\textbf{\textit{Learning Rate and Epochs.}} We report the impact of learning rate and epochs on our method performance in Table~\ref{tab: lr-epoch-effect}. Across both variants of the method, higher learning rates and epochs consistently resulted in a larger number of valid CF examples found (i.e. higher success rate) and higher explanation size. This behavior is expected: larger learning rates and a higher number of epochs allow the perturbation process to move farther from the original hypergraph, increasing the probability of crossing the model decision boundary, but typically at the cost of less sparse structural interventions.

\begin{table}[t]
\centering
\caption{Effect of learning rate and optimization epochs on counterfactual success rate and explanation size on Cora-CA dataset.}
\label{tab: lr-epoch-effect}
{\scriptsize
\setlength{\tabcolsep}{2.8pt}
\renewcommand{\arraystretch}{0.88}
\resizebox{\columnwidth}{!}{%
\begin{tabular}{lllccc}
\hline
\textbf{Method} & \textbf{Metric} & \textbf{Learning Rate} & \multicolumn{3}{c}{\textbf{Epochs}} \\
\cline{4-6}
& & & \textbf{200} & \textbf{400} & \textbf{600} \\
\hline
\multirow{6}{*}{\vone}
& \multirow{3}{*}{Success Rate $\uparrow$}
& 1   & 0.098305 & 0.168942 & 0.217617 \\
& & 10  & 0.319643 & 0.419890 & 0.447761 \\
& & 100 & 0.536062 & 0.600000 & 0.626016 \\
\cline{2-6}
& \multirow{3}{*}{Explanation Size $\downarrow$}
& 1   & 0.116949 & 0.215017 & 0.278066 \\
& & 10  & 0.441071 & 0.570902 & 0.611940 \\
& & 100 & 0.750487 & 0.862626 & 0.914634 \\
\hline
\multirow{6}{*}{\vthree}
& \multirow{3}{*}{Success Rate $\uparrow$}
& 1   & 0.088136 & 0.140678 & 0.162712 \\
& & 10  & 0.245675 & 0.309028 & 0.349123 \\
& & 100 & 0.446847 & 0.500921 & 0.537594 \\
\cline{2-6}
& \multirow{3}{*}{Explanation Size $\downarrow$}
& 1   & 0.130508 & 0.255932 & 0.371186 \\
& & 10  & 0.851211 & 1.128470 & 1.345610 \\
& & 100 & 1.266670 & 1.489870 & 1.597740 \\
\hline
\end{tabular}%
}}
\end{table}

\noindent \textbf{\textit{Adaptive Learning Rate Strategy.}} Due to the impossibility of having a one-fit-all learning rate value, we use a dynamic learning rate adaptation strategy in three variants, as described in Section~\ref{subsec: hyperparameter-search}. The comparison between these strategies is intended as an ablation rather than part of the full benchmark evaluation, therefore we report it on two representative datasets, \texttt{Cora-CA} and \texttt{Mushroom}, and omit the full-dataset table due to space constraints. We empirically observe that, when using \texttt{EE} as the learning rate adaptation strategy, the learning rate starts at a higher value and gradually decreases throughout the CF generation process. This observation is consistent with the results reported in Table~\ref{tab: lr-adaptive-strategy-comparison}. In particular, \texttt{1E} sets the learning rate at the first epoch only; given the observed decay, this value is larger than the learning rates that would be selected by \texttt{Po2} or \texttt{EE} at later epochs. As a result, \texttt{1E} achieves a higher success rate, but at the cost of a larger explanation size.

Table~\ref{tab: lr-adaptive-strategy-comparison} also shows that using \texttt{EE} is not beneficial, since the slowdown it introduces is not compensated by improvements in either success rate or explanation size. Moreover, for \vone{}, the use of \texttt{Po2} introduces a slowdown of approximately $200\%$, while reducing the explanation size by only approximately $3.4\%$ on the \texttt{Cora-CA} dataset and yielding no reduction on the \texttt{Mushroom} dataset. Therefore, for \vone{}, we adopt the \texttt{1E} strategy.

Finally, for \vthree{}, \texttt{Po2} introduces a slowdown of approximately $110\%$, but reduces the explanation size by approximately $36\%$ on the \texttt{Cora-CA} dataset and by approximately $28\%$ on the \texttt{Mushroom} dataset. Therefore, for \vthree{}, we adopt the \texttt{Po2} strategy.

\begin{table}[t]
\centering
\caption{Comparison of adaptive learning-rate strategies on the Cora-CA and Mushroom datasets. The strategies \texttt{1E}, \texttt{Po2}, and \texttt{EE} denote adaptation at the first epoch only, at powers of two epochs, and at every epoch, respectively.}
\label{tab: lr-adaptive-strategy-comparison}
{\footnotesize
\setlength{\tabcolsep}{4.0pt}
\renewcommand{\arraystretch}{0.88}
\resizebox{\columnwidth}{!}{%
\begin{tabular}{ll l cc}
\hline
\textbf{Metric} & \textbf{Method} & \textbf{Strategy} & \textbf{Cora-CA} & \textbf{Mushroom} \\
\hline
\multirow{6}{*}{Success Rate $\uparrow$}
& \vone & 1E  & 1.0000 & 1.0000 \\
& \vone & Po2 & 0.9907 & 1.0000 \\
& \vone & EE  & 0.9595 & 0.9472 \\
& \vthree & 1E  & 0.8915 & 1.0000 \\
& \vthree & Po2 & 0.7991 & 1.0000 \\
& \vthree & EE  & 0.8098 & 0.8712 \\
\hline
\multirow{6}{*}{Explanation Time $\downarrow$}
& \vone & 1E  & 0.0352 & 0.7208 \\
& \vone & Po2 & 0.1119 & 1.9283 \\
& \vone & EE  & 0.5203 & 14.033 \\
& \vthree & 1E  & 0.0968 & 0.9743 \\
& \vthree & Po2 & 0.1910 & 2.1713 \\
& \vthree & EE  & 0.5080 & 14.417 \\
\hline
\multirow{6}{*}{Slow Down $\downarrow$}
& \vone & 1E  & 0.000\% & 0.000\% \\
& \vone & Po2 & 217.8\% & 167.5\% \\
& \vone & EE  & 1377\% & 1847\% \\
& \vthree & 1E  & 0.000\% & 0.000\% \\
& \vthree & Po2 & 97.43\% & 122.9\% \\
& \vthree & EE  & 425.0\% & 1380\% \\
\hline
\multirow{6}{*}{Explanation Size $\downarrow$}
& \vone & 1E  & 1.7059 & 4.5157 \\
& \vone & Po2 & 1.6472 & 4.5173 \\
& \vone & EE  & 1.5743 & 4.2426 \\
& \vthree & 1E  & 2.3256 & 18.582 \\
& \vthree & Po2 & 1.4879 & 13.333 \\
& \vthree & EE  & 1.4808 & 3.9083 \\
\hline
\end{tabular}%
}}
\end{table}

\noindent \textbf{\textit{Regularization Parameter.}} Following original work on counterfactual explanations~\cite{wachter2017hjlt}, we select the largest value of $\beta$ that still ensures a prediction flip, choosing it through an iterative algorithm. This differs from the approach chosen by CF-GNNExplainer~\cite{CF-GNNExplainer} where they fix a $\beta$ for all data points of all datasets. Table~\ref{tab: beta-cf-hypergnnexplainer} reports a comparison between our choice (i.e. Largest) and the choice done by CF-GNNExplainer~\cite{CF-GNNExplainer} with fixed $\beta \in \{0.1, 0.5\}$ for the \vone{} variant of our method. As expected, selecting the largest value of $\beta$ depending on the  data point, allows to hit a higher success rate, as it avoids too hard regularizations when they are counterproductive. The price for our choice, is a higher time to produce a CF explanation, as this now includes the time needed for selecting $\beta$. Notably, on the \texttt{House} dataset, fixing $\beta=0.5$ prevents the \vone{} variant from generating any counterfactual explanations. Due to space constraints, we omit the corresponding table for the \vthree{} variant; however, the observed trends are consistent and lead to the same conclusion. A further observation from Table~\ref{tab: beta-cf-hypergnnexplainer} is that the adaptive selection of $\beta$ does not necessarily yield a success rate equal to one. Although $\beta$ is chosen as the largest value that induces a prediction flip during the optimization procedure, this flip may not always be obtained. This can occur for two main reasons. First, for some instances, the method may fail to identify a valid counterfactual even in the absence of regularization, i.e., when $\beta=0$. Second, a prediction flip obtained during optimization using the continuous soft mask may be lost after thresholding the mask to obtain the final binary incidence matrix.

\begin{table*}[t]
\centering
\caption{Comparison of regularization parameter selection strategies across datasets. Fixed values $\beta = 0.1$ and $\beta = 0.5$ are evaluated against the instance specific largest value of $\beta$ that preserves success in counterfactual generation.}
\label{tab: beta-cf-hypergnnexplainer}
{\scriptsize
\setlength{\tabcolsep}{3.0pt}
\renewcommand{\arraystretch}{0.90}
\resizebox{\textwidth}{!}{%
\begin{tabular}{llcccccccccc}
\hline
\textbf{Metric} & \textbf{Beta} 
& \textbf{Cora} 
& \textbf{Citeseer} 
& \textbf{Pubmed} 
& \textbf{Cora-CA} 
& \textbf{DBLP-CA} 
& \textbf{Zoo} 
& \textbf{Mushroom} 
& \textbf{NTU2012} 
& \textbf{ModelNet40} 
& \textbf{House} \\
\hline
\multirow{3}{*}{Success Rate $\uparrow$}
& 0.1 & 0.96 & 0.91 & 0.92 & 0.93 & 0.92 & 1.00 & 0.98 & 0.97 & 0.93 & 0.54 \\
& 0.5 & 0.96 & 0.91 & 0.79 & 0.89 & 0.86 & 1.00 & 0.90 & 0.93 & 0.83 & NA \\
& Largest & 0.96 & 0.97 & 0.98 & 0.98 & 0.96 & 1.00 & 0.99 & 0.97 & 1.00 & 0.24  \\
\hline
\multirow{3}{*}{Time $\downarrow$}
& 0.1 & 0.11 & 0.10 & 0.20 & 0.08 & 0.14 & 0.15 & 1.41 & 0.11 & 0.11 & 0.40 \\
& 0.5 & 0.16 & 0.15 & 0.34 & 0.11 & 0.19 & 0.20 & 2.23 & 0.16 & 0.18 & NA \\
& Largest & 1.14 & 1.18 & 2.96 & 2.10 & 1.41 & 2.69 & 14.2 & 1.99 & 1.58 & 2.17  \\
\hline
\multirow{3}{*}{Expl. Size $\downarrow$}
& 0.1 & 2.16 & 1.70 & 5.58 & 1.31 & 1.81 & 9.91 & 3.02 & 3.17 & 2.46 & 2.00 \\
& 0.5 & 3.16 & 1.96 & 3.74 & 1.37 & 1.37 & 9.71 & 2.62 & 2.91 & 2.37 & NA \\
& Largest & 2.41 & 1.76 & 7.44 & 1.64 & 2.21 & 10.63 & 4.50 & 4.02 & 4.24 & 4.08  \\
\hline
\multirow{3}{*}{Sparsity $\uparrow$}
& 0.1 & 0.99 & 0.95 & 1.00 & 0.98 & 0.99 & 0.99 & 1.00 & 1.00 & 1.00 & 1.00 \\
& 0.5 & 1.00 & 0.97 & 1.00 & 0.98 & 0.99 & 0.99 & 1.00 & 1.00 & 1.00 & NA \\
& Largest & 0.98 & 0.89 & 0.99 & 0.95 & 0.97 & 0.99 & 1.00 & 0.99 & 0.99 & 1.00 \\
\hline
\end{tabular}%
}}
\end{table*}

\subsection{Comparison with Graph-Based Methods}
Table~\ref{tab: cf-comparison} shows the performance difference between our method and the graph-based baseline CF-GNNExplainer on the benchmark datasets, including explanation time and speed-up, but excluding the time required to convert the hypergraph into a graph. For a fair comparison, we did not use the largest $\beta$ that would have result in a label flip, but instead fixed $\beta=0.5$ as CF-GNNExplainer does. This does not reward our method's success rate nor penalize its explanation time. A first observation is that for \texttt{DBLP-CA} and \texttt{House} datasets, we were unable to run CF-GNNExplainer, as the star expansion of an hypergraph with incidence matrix $H\in \mathbb{R}^{N \times M}$ produces a graph with adjacency matrix $A \in \mathbb{R}^{(N+M) \times (N+M)}$, which is substantially larger and fails to fit the memory. On the remaining datasets, both variants of our method outperform the graph-native baseline on success rate and explanation size; this demonstrates that our hypergraph-native approach is more effective at producing counterfactual explanation for hypergraph data. The same table reports also the average time taken to generate a counterfactual explanation, where we can see that both variants of our method are generally faster than CF-GNNExplainer. We attribute these gains in success rate and explanation size to the representational advantage of hypergraphs over standard graphs. Hypergraphs naturally encode higher-order interactions, so a single modification in the hypergraph structure may induce a broader effect on the model prediction. Conversely, when the same data are represented as a graph, an analogous semantic change may require multiple edge-level perturbations. This can make graph-based counterfactual explanations larger and less effective in inducing the desired prediction flip.

\begin{table*}[t]
\centering
\caption{Performance comparison between CF-HyperGNNExplainer variants and the graph-based CF-GNNExplainer baseline across benchmark datasets. NA is used whwen no valida counterfactual example has been generated, while ``-'' is used for simulation which could not run.}
\label{tab: cf-comparison}
{\scriptsize
\setlength{\tabcolsep}{3.0pt}
\renewcommand{\arraystretch}{0.92}
\resizebox{\textwidth}{!}{%
\begin{tabular}{llccccccccccc}
\hline
\textbf{Metric} & \textbf{Method} & \textbf{Cora} & \textbf{Citeseer} & \textbf{Pubmed} & \textbf{Cora-CA} & \textbf{DBLP-CA} & \textbf{Zoo} & \textbf{Mushroom} & \textbf{NTU2012} & \textbf{ModelNet40} & \textbf{House} \\
\hline
\multirow{3}{*}{Success Rate $\uparrow$}
& CF-GNNExplainer & 0.28 & 0.20 & 0.45 & 0.38 & -- & 0.53 & 0.36 & 0.41 & 0.38 & -- \\
& Ours (NHP) & 0.94 & 1.00 & 0.79 & 0.88 & 0.88 & 1.00 & 0.96 & 0.93 & 0.83 & NA \\
& Ours (HP) & 0.91 & 0.82 & 0.90 & 0.80 & 0.85 & 1.00 & 1.00 & 0.97 & 0.95 & 0.40 \\
\hline
\multirow{3}{*}{Expl. Size $\downarrow$}
& CF-GNNExplainer & 22.1 & 9.74 & 13.6 & 3.53 & -- & 5.56 & 1.2 & 6.89 & 9.07 & -- \\
& Ours (NHP) & 3.16 & 1.96 & 3.74 & 1.37 & 1.37 & 9.71 & 2.62 & 2.91 & 2.37 & NA \\
& Ours (HP) & 1.56 & 1.41 & 4.71 & 1.55 & 2.21 & 8.35 & 22.79 & 4.32 & 4.16 & 54.12 \\
\hline
\multirow{3}{*}{Sparsity $\uparrow$}
& CF-GNNExplainer & 0.81 & 0.63 & 0.99 & 0.88 & -- & 1.00 & 1.00 & 0.94 & 0.94 & --  \\
& Ours (NHP) & 1.00 & 0.97 & 1.00 & 0.98 & 0.99 & 0.99 & 1.00 & 1.00 & 1.00 & NA \\
& Ours (HP) & 0.99 & 0.90 & 1.00 & 0.92 & 0.97 & 0.86 & 0.97 & 0.98 & 0.98 & 0.99 \\
\hline
\multirow{3}{*}{Time $\downarrow$}
& CF-GNNExplainer & 0.30 & 0.28 & 0.73 & 0.68 & -- & 0.58 & 1.74 & 1.32 & 1.13 & -- \\
& Ours (NHP) & 0.16 & 0.15 & 0.34 & 0.11 & 0.19 & 0.20 & 2.23 & 0.16 & 0.18 & NA \\
& Ours (HP) & 0.17 & 0.15 & 0.42 & 0.19 & 0.23 & 0.18 & 2.19 & 0.19 & 0.15 & 0.36 \\
\hline
\multirow{3}{*}{Speedup $\uparrow$}
& CF-GNNExplainer & 1.00x & 1.00x & 1.00x & 1.00x & NA & 1.00x & 1.00x & 1.00x & 1.00x & NA \\
& Ours (NHP) & 1.88x & 1.87x & 2.15x & 6.18x & NA & 2.90x & 0.78x & 8.25x & 6.28x & NA \\
& Ours (HP) & 1.76x & 1.87x & 1.74x & 3.58x & NA & 3.22x & 0.80x & 6.95x & 7.53x & NA \\
\hline
\end{tabular}%
}}
\end{table*}

\subsection{Comparison with Hypergraph-Based Methods}

Table~\ref{tab: hypergraphs-method-comparison} shows results obtained by comparing our method variants with HyperEX~\cite{HyperEX} and SHypX~\cite{shypx}. Before discussing the comparison, it is essential to keep in mind that our method falls within the XAI subfield of counterfactaul explainability, hence it does not aim to outperform existing methods across all metrics, but rather aims to achieve the highest label flipping success rate ($\mathrm{Fid}_{+}^{\mathrm{acc}}$) with the minimal intervention on the hypergraph topology ($\mathrm{Expl.\,\, Size}$). More broadly, counterfactual explainability focuses on the prediction on the counterfactual explanation rather than the counterfactual distance, and hence focuses on the metrics computed on the counterfactual hypergraph, denoted by $\mathrm{Fid}_{+}$, not on the metrics computed on the removed substructure, denoted by $\mathrm{Fid}_{-}$. This is a fundamental point, especially with respect to SHypX which does exactly the opposite. Table~\ref{tab: hypergraphs-method-comparison} reports the mean value of each metric computed on the test set instances; standard deviations are omitted due to space contraints. We observe that both variants of our method succeed in the main objective of our method: achieving high $\mathrm{Fid}_{+}^{\mathrm{acc}}$ while simultaneously maintaining a small explanation size. More generally, both variants achieves high $\mathrm{Fid}_{+}^{\mathrm{acc}}$, $\mathrm{Fid}_{+}^{\mathrm{KL}}$, and $\mathrm{Fid}_{+}^{\mathrm{TV}}$, showing that the generated explanations successfully alter both the predicted class and the predictive distribution. However, the cross-entropy-based metric $\mathrm{Fid}_{+}^{\mathrm{xent}}$ behaves differently as it combines distributional disagreement with the entropy of the counterfactual prediction. Lower $\mathrm{Fid}_{+}^{\mathrm{xent}}$ values, combined with the other metrics, therefore do not necessarily indicate weaker counterfactual explanations; rather, they suggest that our explanations induce less extreme or less uncertain predictive distributions than competing methods. 
Concerning the $\mathrm{Fid}_{-}$ metrics, the results should be interpreted with care. These metrics measure whether the removed substructure, i.e., $H-H'$, is by itself sufficient to reproduce the original prediction, and are therefore closer to a notion of factual sufficiency than to counterfactual validity. This explains why SHypX obtains the best values across almost all $\mathrm{Fid}_{-}$ metrics: SHypX is explicitly designed to extract a faithful subhypergraph that preserves the original prediction, whereas our method is designed to remove the structural evidence needed to change it. Consequently, low $\mathrm{Fid}_{-}$ values are aligned with the objective of SHypX, but they are not the primary desideratum of a counterfactual explainer.

Nevertheless, the comparison on the $\mathrm{Fid}_{-}$ metrics provides useful diagnostic information. The HP variant achieves substantially lower $\mathrm{Fid}_{-}$ values than NHP on most datasets and metrics. This is expected, since HP removes entire hyperedges and therefore $H-H'$ remains a more coherent hypergraph fragment. In contrast, NHP removes individual node-hyperedge incidences, so the counterfactual difference may consist of sparse or fragmented incidence patterns that are not sufficient, in isolation, to reproduce the original prediction. This explains why NHP is usually stronger on the $\mathrm{Fid}_{+}$ metrics, especially $\mathrm{Fid}_{+}^{\mathrm{acc}}$, but weaker on the $\mathrm{Fid}_{-}$ metrics.

Overall, the $\mathrm{Fid}_{-}$ metrics confirm the methodological difference between factual and counterfactual explainers. SHypX is superior when the goal is to find a compact subhypergraph that preserves the original decision. Our method, instead, is superior on the metrics that directly evaluate counterfactual behavior, namely whether the prediction changes after the intervention and whether this is achieved with a small edit. Therefore, the lower performance of our method on $\mathrm{Fid}_{-}$ should not be interpreted as a failure, but rather as a consequence of optimizing for a different explanation objective.

Finally, our variants are slower than SHypX because they solve a per-instance counterfactual optimization problem and include the additional overhead of selecting the largest $\beta$ that still produces a valid prediction flip. The runtime of HyperEX is not reported because the comparison would not be directly meaningful: HyperEX shifts most of the cost to a precomputation stage, which is expensive when only one explanation is required but becomes negligible per instance once amortized over many explanations.

\begin{table*}[t]
\centering
\caption{Performance comparison between CF-HyperGNNExplainer variants and the hypergraph-native baselines across benchmark datasets.}
\label{tab: hypergraphs-method-comparison}
\resizebox{\textwidth}{!}{%
\begin{tabular}{llccccccccccc}
\hline
\textbf{Metric} & \textbf{Model} 
& \textbf{Cora} 
& \textbf{Citeseer} 
& \textbf{Pubmed} 
& \textbf{Cora-CA} 
& \textbf{DBLP-CA} 
& \textbf{Zoo} 
& \textbf{Mushroom} 
& \textbf{NTU2012} 
& \textbf{ModelNet40} 
& \textbf{House} \\
\hline

\multirow{4}{*}{$\mathrm{Fid}_{-}^{\mathrm{acc}} \downarrow$}
& HyperEX    & 0.25 & 0.13 & 0.23 & 0.22 & 0.18 & 0.63 & 0.53 & 0.32 & 0.24 & 0.16 \\
& SHypX      & 0.00 & 0.01 & 0.02 & 0.00 & 0.01 & 0.05 & 0.00 & 0.01 & 0.00 & 0.04 \\
& Ours (NHP) & 0.28 & 0.18 & 0.26 & 0.30 & 0.20 & 0.23 & 0.28 & 0.35 & 0.24 & 0.16 \\
& Ours (HP)  & 0.02 & 0.03 & 0.01 & 0.03 & 0.03 & 0.12 & 0.00 & 0.12 & 0.06 & 0.02 \\
\hline

\multirow{4}{*}{$\mathrm{Fid}_{+}^{\mathrm{acc}} \uparrow$}
& HyperEX    & 0.57 & 0.76 & 0.51 & 0.70 & 0.81 & 0.00 & 0.00 & 0.93 & 0.89 & 0.04 \\
& SHypX      & 0.44 & 0.53 & 0.17 & 0.49 & 0.42 & 0.21 & 0.14 & 0.36 & 0.29 & 0.00 \\
& Ours (NHP) & 0.96 & 0.97 & 0.98 & 0.98 & 0.96 & 1.00 & 0.99 & 0.97 & 1.00 & 0.24 \\
& Ours (HP)  & 0.73 & 0.80 & 0.92 & 0.77 & 0.75 & 1.00 & 0.99 & 0.87 & 0.95 & 0.52 \\
\hline

\multirow{4}{*}{$\mathrm{Fid}_{-}^{\mathrm{KL}} \downarrow$}
& HyperEX    & 2.09 & 1.53 & 0.75 & 1.71 & 0.83 & 8.93 & 8.09 & 1.90 & 1.89 & 0.00 \\
& SHypX      & 0.00 & 0.00 & 0.01 & 0.00 & 0.00 & 0.03 & 0.00 & 0.01 & 0.01 & 0.00 \\
& Ours (NHP) & 2.17 & 0.93 & 0.83 & 2.31 & 1.00 & 1.50 & 3.80 & 2.15 & 1.84 & 0.00 \\
& Ours (HP)  & 0.15 & 0.08 & 0.12 & 0.14 & 0.12 & 0.70 & 0.04 & 0.66 & 0.16 & 0.00 \\
\hline

\multirow{4}{*}{$\mathrm{Fid}_{+}^{\mathrm{KL}} \uparrow$}
& HyperEX    & 8.26 & 6.79 & 1.11 & 9.33 & 8.07 & 0.00 & 0.00 & 13.4 & 11.3 & 0.00 \\
& SHypX      & 4.51 & 4.35 & 0.31 & 5.08 & 3.07 & 2.82 & 1.67 & 3.01 & 2.06 & 0.00 \\
& Ours (NHP) & 7.45 & 7.94 & 1.74 & 7.99 & 8.04 & 14.0 & 14.4 & 10.7 & 10.6 & 0.00 \\
& Ours (HP)  & 5.45 & 5.53 & 1.32 & 5.57 & 5.90 & 6.67 & 13.4 & 8.59 & 9.62 & 0.00 \\
\hline

\multirow{4}{*}{$\mathrm{Fid}_{-}^{\mathrm{TV}} \downarrow$}
& HyperEX    & 0.26 & 0.23 & 0.27 & 0.22 & 0.19 & 0.67 & 0.53 & 0.32 & 0.26 & 0.00 \\
& SHypX      & 0.01 & 0.01 & 0.03 & 0.01 & 0.01 & 0.04 & 0.00 & 0.02 & 0.01 & 0.01 \\
& Ours (NHP) & 0.29 & 0.18 & 0.29 & 0.30 & 0.23 & 0.24 & 0.28 & 0.35 & 0.26 & 0.00 \\
& Ours (HP)  & 0.07 & 0.05 & 0.11 & 0.06 & 0.06 & 0.13 & 0.01 & 0.17 & 0.08 & 0.00 \\
\hline

\multirow{4}{*}{$\mathrm{Fid}_{+}^{\mathrm{TV}} \uparrow$}
& HyperEX    & 0.65 & 0.64 & 0.39 & 0.77 & 0.79 & 0.00 & 0.00 & 0.91 & 0.89 & 0.01 \\
& SHypX      & 0.45 & 0.51 & 0.14 & 0.52 & 0.40 & 0.32 & 0.14 & 0.35 & 0.28 & 0.00 \\
& Ours (NHP) & 0.80 & 0.76 & 0.55 & 0.81 & 0.82 & 0.88 & 0.96 & 0.83 & 0.89 & 0.00 \\
& Ours (HP)  & 0.58 & 0.60 & 0.48 & 0.61 & 0.62 & 0.79 & 0.94 & 0.70 & 0.83 & 0.00 \\
\hline

\multirow{4}{*}{$\mathrm{Fid}_{-}^{\mathrm{Xent}} \downarrow$}
& HyperEX    & 1.93 & 2.39 & 0.94 & 1.86 & 1.07 & 9.93 & 8.14 & 2.27 & 2.11 & 0.69 \\
& SHypX      & 0.13 & 0.18 & 0.27 & 0.14 & 0.22 & 0.17 & 0.01 & 0.45 & 0.20 & 0.69 \\
& Ours (NHP) & 2.45 & 0.93 & 1.35 & 2.79 & 1.30 & 4.23 & 5.10 & 3.07 & 2.01 & 0.69 \\
& Ours (HP)  & 0.59 & 0.36 & 0.82 & 0.44 & 0.44 & 2.89 & 0.06 & 1.40 & 0.62 & 0.69 \\
\hline

\multirow{4}{*}{$\mathrm{Fid}_{+}^{\mathrm{Xent}} \uparrow$}
& HyperEX    & 9.75 & 8.27 & 1.95 & 11.1 & 8.91 & 0.20 & 0.00 & 16.3 & 14.3 & 0.69 \\
& SHypX      & 5.32 & 5.67 & 0.89 & 6.20 & 3.55 & 3.34 & 1.74 & 4.18 & 3.02 & 0.69 \\
& Ours (NHP) & 2.78 & 2.12 & 1.19 & 2.60 & 3.08 & 6.03 & 4.64 & 4.69 & 3.91 & 0.69 \\
& Ours (HP)  & 2.00 & 1.70 & 1.04 & 2.10 & 2.35 & 2.91 & 5.31 & 3.87 & 3.59 & 0.69 \\
\hline

\multirow{4}{*}{$\mathrm{Time} \downarrow$}
& HyperEX    & -- & -- & -- & -- & -- & -- & -- & -- & -- & -- \\
& SHypX      & 0.24 & 0.20 & 1.09 & 0.43 & 1.03 & 0.51 & 3.77 & 0.50 & 0.50 & 0.50 \\
& Ours (NHP) & 1.14 & 1.18 & 2.96 & 2.10 & 1.41 & 2.69 & 14.2 & 1.99 & 1.58 & 2.17 \\
& Ours (HP)  & 2.18 & 2.35 & 8.02 & 2.95 & 3.21 & 4.41 & 27.6 & 4.63 & 3.43 & 7.87 \\
\hline

\multirow{4}{*}{$\mathrm{Expl.\,\, Size} \downarrow$}
& HyperEX    & 9.70 & 8.69 & 9.77 & 9.16 & 9.48 & 10.0 & 10.0 & 10.0 & 10.0 & 10.0 \\
& SHypX      & 5.82 & 3.68 & 14.02 & 4.48 & 5.05 & 35.21 & 8.33 & 12.8 & 8.89 & 0.00 \\
& Ours (NHP) & 2.41 & 1.76 & 7.44 & 1.64 & 2.21 & 10.63 & 4.50 & 4.02 & 4.24 & 4.08 \\
& Ours (HP)  & 1.56 & 1.41 & 4.71 & 1.55 & 2.21 & 8.35 & 22.79 & 4.32 & 4.16 & 54.12 \\
\hline

\multirow{4}{*}{$\mathrm{Density} \downarrow$}
& HyperEX    & 0.10 & 0.34 & 0.03 & 0.28 & 0.16 & 0.01 & 0.00 & 0.03 & 0.02 & 0.00 \\
& SHypX      & 0.03 & 0.15 & 0.01 & 0.12 & 0.07 & 0.00 & 0.02 & 0.02 & 0.01 & 0.00 \\
& Ours (NHP) & 0.02 & 0.11 & 0.01 & 0.05 & 0.03 & 0.01 & 0.00 & 0.01 & 0.01 & 0.00 \\
& Ours (HP)  & 0.02 & 0.11 & 0.00 & 0.11 & 0.05 & 0.19 & 0.08 & 0.06 & 0.04 & 0.16 \\
\hline

\end{tabular}%
}
\end{table*}

\section{Limitations}
\textbf{\textit{Sparse Hypergraphs.}}
The \vone{} variant works by removing node-hyperedge incidences. For a node $\nodev$ with degree $d_\nodev$, there exist $2^{d_\nodev}$ possible combinations of incidences that can be kept or removed. When $d_\nodev$ is small enough, exhaustive enumeration of all possible combinations becomes feasible and more efficient than the proposed method.

This limitation is specific to the \vone{} variant, due to the node-centric perturbation strategy. The \vthree{} strategy, which operates on hyperedges within n-hop neighborhood, typically generates a larger search space and therefore benefits more from gradient-based optimization. 

\noindent \textbf{\textit{Random Baseline Performance and Theoretical Lower Bound.}}
Random perturbation baselines are able to find a large number of valid counterfactuals, but typically generate explanations with significantly larger explanation sizes compared to the gradient-based approaches. However, for the \vone{} variant in very sparse hypergraphs, there exists a lower bound on the probability of randomly discovering the smallest counterfactual.

Assume that the smallest counterfactual exists and is unique. For a target node $\nodev$ with degree $d_\nodev$, the probability of randomly obtaining the minimal counterfactual in $t$ independent sampling attempts, where each incidence is kept or removed with probability $\frac{1}{2}$, is given by:
\begin{equation}
\label{eq:lower_bound}
P(d_\nodev, t) = 1 - \bigg( 1 - \bigg(\frac{1}{2}\bigg)^{d_\nodev} \bigg )^{t}
\end{equation}
This probability follows from the theory of Bernoulli trials, where each sampling attempt represents an independent trial with success probability $\left(\frac{1}{2}\right)^{d_\nodev}$. It is important to note that equation~\eqref{eq:lower_bound} represents a theoretical lower bound on the probability, as in practice there may exist multiple counterfactuals at the same minimal distance from the original sample.

This lower bound demonstrates that in very sparse hypergraphs, the probability of randomly generating the minimal counterfactual becomes significantly high, even with a small number of sampling attempts. For example, with $d_\nodev = 5$ and $t=100$ attempts, the probability exceeds $0.95$.

\noindent \textbf{\textit{Restricted Counterfactual Search Space.}} A further limitation of CF-HyperGNNExplainer is that the current counterfactual search space is restricted to deletion operations on the hypergraph topology. As a consequence, the method can only identify counterfactuals that are reachable through structural sparsification of the original hypergraph. In our experiments, we observed that for some instances no valid counterfactual explanation could be generated, suggesting that a prediction flip may require interventions outside this deletion-only search space. Extending the method to richer interventions is, however, non-trivial. Allowing perturbations of node features would make the counterfactual generation process no longer bounded solely to the hypergraph topology, thereby changing the nature of the explanation from a purely structural one to a joint structural-feature explanation. Conversely, allowing additions of node–hyperedge incidences or hyperedges would substantially enlarge the search space: adding an incidence or a new hyperedge may expand the n-hop neighborhood of the target node and introduce new information-flow paths that are not constrained by the original local neighborhood. Designing principled and computationally tractable counterfactual objectives that account for such feature-level or additive structural interventions is therefore left as future work.

\section{Conclusions and Future Work}
\label{sec: conclusions}

In this work, we introduced CF-HyperGNNExplainer, the first counterfactual explanation method specifically designed for Hypergraph Neural Networks. Our approach generates counterfactual explanations by performing minimal edits to the hypergraph incidence matrix, either by removing node-hyperedge incidences (\vone{} variant) or by removing entire hyperedges (\vthree{} variant).

Experimental results on multiple hypergraph benchmark datasets show that CF-HyperGNNExplainer generates valid and concise counterfactual explanations. Both \vone{} and \vthree{} achieve high label flipping success rates while requiring only small structural edits, confirming that targeted node-hyperedge or hyperedge level perturbations can effectively alter HGNN predictions. Compared with CF-GNNExplainer applied to star-expanded hypergraphs, our hypergraph-native formulation yields higher success rates, smaller explanations, and generally faster generation, while avoiding the scalability issues caused by graph expansion on larger datasets. Comparisons with HyperEX and SHypX further clarify that, unlike factual explainers that seek compact sub-hypergraphs preserving the original decision, CF-HyperGNNExplainer is optimized to change the decision through minimal interventions.

Both variants of CF-HyperGNNExplainer are currently restricted to deletion operations on the incidence matrix and are designed for node classification. Future work will broaden the set of allowable interventions to generate richer counterfactuals, including perturbations of node features and the addition of incidences or hyperedges. Another promising direction is to extend the method to hypergraph-level prediction tasks (hypergraph classification), which, while less common than node classification, are increasingly relevant in several application domains.

  



\FloatBarrier
\bibliographystyle{ACM-Reference-Format}
\bibliography{bibliography}

\end{document}